%% file: root.tex
\documentclass[a4paper]{styles/svproc}

\usepackage{url}
\usepackage{type1cm}
\usepackage{subeqnarray}
\usepackage{footmisc}
\usepackage{graphicx}
\usepackage{xspace}
\usepackage{enumitem}

\input{custom/packages}

\input{custom/macros}

\begin{document}
\mainmatter

\title{Scaling Long-Horizon Online POMDP Planning via Rapid State Space Sampling}

\titlerunning{Online POMDP Planning via Rapid State Space Sampling}

\author{Yuanchu Liang\inst{1}\footnote[1]{Authors have equal contributions.} \and Edward Kim\inst{1}\footnotemark[1]{} \and Wil Thomason\inst{2}\footnotemark[1]{} \and 
Zachary Kingston\inst{2}\footnotemark[1]{} \and Hanna Kurniawati\inst{1} \and Lydia E. Kavraki\inst{2}}

\authorrunning{Liang et al.}
\tocauthor{Yuanchu Liang, Edward Kim, Zachary Kingston, Wil Thomason,
Hanna Kurniawati and Lydia E. Kavraki}

\institute{Australian National University, Canberra ACT 2601, Australia,\\
\email{\{Yuanchu.Liang, Edward.Kim, hanna.kurniawati\}@anu.edu.au}
\and
Rice University, Houston TX 77005, USA,\\
\email{\{wbthomason, zak, kavraki\}@rice.edu}}

\maketitle

\begin{abstract}
Partially Observable Markov Decision Processes (\pomdps) are a general and principled framework for motion planning under uncertainty. Despite tremendous improvement in the scalability of \pomdp solvers, long-horizon \pomdps (\eg{} $\geq15$ steps) remain difficult to solve. This paper proposes a new approximate online \pomdp solver, called \nopLong (\nop). \nop uses novel extremely fast sampling-based motion planning techniques to sample the state space and generate a diverse set of macro actions online which are then used to bias belief-space sampling and infer high-quality policies \emph{without} requiring exhaustive enumeration of the action space---a fundamental constraint for modern online \pomdp solvers. \nop is evaluated on various long-horizon \pomdps, including on a problem with a planning horizon of more than 100 steps and a problem with a 15-dimensional state space that requires more than 20 look ahead steps. In all of these problems, \nop \emph{substantially outperforms} other state-of-the-art methods by up to \emph{multiple folds}.
\end{abstract}
\section{Introduction}

\input{sections/introduction}

\section{Background and Related Work}

\subsection{Sampling-Based Motion Planners}

\input{sections/sbmp}

\subsection{POMDP Background}

\input{sections/pomdp}

\subsection{Long-Horizon POMDPs}\label{sec: long horizon POMDPs}
\input{sections/largepomdps}

\section{POMDP Planning with SBMP-Generated Trajectories}

\input{sections/integration-james}

\section{\nopTitle}
\label{sec.nop}

\input{sections/pop-art}

\section{Experiments} \label{sec:experiments}
\input{sections/experiments}

\section{Summary} \label{sec: Discussion}
\input{sections/discussion}

\section{Acknowledgements}
YL, EK, and HK have been partially supported by the ANU Futures Scheme. YL has been supported by Australia RTP scholarship. HK has been partially supported by the SmartSat CRC. ZK, WT and LEK have been supported in part by NSF 2008720, 2336612, and Rice University Funds. WT has been supported by NSF ITR 2127309---CRA CIFellows Project.

\bibliographystyle{abbrv}
\bibliography{references}

\end{document}

%% file: custom/packages.tex
\usepackage[T1]{fontenc}%
\usepackage[utf8]{inputenc}%

\usepackage{xcolor}
\usepackage{amsmath} %
\usepackage{mathrsfs} %
\usepackage{amssymb} %
\usepackage{algorithm} %
\usepackage{algpseudocode} %
\usepackage{subcaption}
\usepackage{mwe}
\usepackage{graphicx}
\usepackage{multirow}
\usepackage{tablefootnote}

\usepackage{mathtools}
\mathtoolsset{showonlyrefs}

\usepackage[
	activate   = {true},
	protrusion = false,
	expansion  = true,
	kerning    = true,
	spacing    = true,
	tracking   = false,
	auto       = true,
	selected   = true,
	factor     = 2000,
	stretch    = 20,
	shrink     = 20,
]{microtype}

\definecolor{mycolor}{HTML}{BE830E} %

\usepackage[pdfa,colorlinks,bookmarksopen,bookmarksnumbered,allcolors=mycolor]{hyperref}
\usepackage[english]{babel}

%% file: custom/macros.tex
\newenvironment{myDescription}
{\begin{description}[leftmargin=0pt, labelindent=-5pt]
  \setlength{\leftskip}{0pt}
  \setlength{\labelsep}{0pt}
  \setlength{\itemsep}{0pt}
  \setlength{\parskip}{3pt}
  \setlength{\parsep}{0pt}
  \setlength{\partopsep}{0pt}}
{\end{description}}

\newenvironment{myItemize}
{\begin{itemize}
  \setlength{\leftskip}{-10pt}
  \setlength{\labelsep}{0pt}
  \setlength{\itemsep}{0pt}
  \setlength{\parskip}{0pt}
  \setlength{\parsep}{0pt}
  \setlength{\partopsep}{0pt}}
{\end{itemize}}

\newcommand{\fref}[1]{Figure~\ref{#1}}
\newcommand{\tref}[1]{Table~\ref{#1}}

\newcommand{\pomcp}{\textsc{pomcp}\xspace}
\newcommand{\despot}{\textsc{despot}\xspace}
\newcommand{\magic}{\textsc{magic}\xspace}
\newcommand{\rmag}{\textsc{rmag}\xspace}
\newcommand{\mdp}{\textsc{mdp}\xspace}
\newcommand{\rpomcp}{\textsc{r-pomcp}\xspace}
\newcommand{\pomdp}{\textsc{pomdp}\xspace}
\newcommand{\pomdps}{\textsc{pomdp}s\xspace}

\newcommand{\kl}{\textsc{kl}}
\newcommand{\samplingHeuristic}{\textsc{SampleHeuristics}\xspace}

\newcommand{\nopLong}{Reference-Based Online \pomdp Planning via Rapid State Space Sampling\xspace} 
\newcommand{\nop}{\textsc{rop-r\scalebox{.8}{a}s\scalebox{.9}{3}}\xspace} 
\newcommand{\nopTitle}{ROP-RaS3\xspace} 

\newcommand{\Reals}{\mathbb{R}} %
\newcommand{\Exp}{\mathbb{E}} %
\newcommand{\Prob}{P} %
\newcommand{\KL}{\operatornamewithlimits{KL}} %
\newcommand{\bigO}{\mathcal{O}} %
\newcommand{\simplex}{\varDelta} %

\newcommand{\States}{\mathcal{S}} %
\newcommand{\Actions}{\mathcal{A}} %
\newcommand{\Observations}{\mathcal{O}} %
\newcommand{\TP}{\mathcal{T}} %
\newcommand{\OP}{\mathcal{Z}} %
\newcommand{\Reward}{R} %
\newcommand{\st}{\ensuremath{s}} %
\newcommand{\nst}{\ensuremath{{s'}}} %
\newcommand{\act}{\ensuremath{a}} %
\newcommand{\mact}{\ensuremath{\vec{\act}}} %
\newcommand{\obs}{\ensuremath{o}} %
\newcommand{\mobs}{\ensuremath{\vec{\obs}}} %
\newcommand{\discount}{{\gamma}} %
\newcommand{\bel}{\ensuremath{b}} %
\newcommand{\belp}{\ensuremath{b'}} %
\newcommand{\hist}{\ensuremath{h}} %
\newcommand{\belSpace}{\ensuremath{\mathcal{B}}} %
\newcommand{\policies}{\ensuremath{\Pi}} %
\newcommand{\Val}{\ensuremath{V}} %
\newcommand{\pol}{{\pi}} %
\newcommand{\optPol}{{\pol}^*} %
\newcommand{\belInit}{\ensuremath{b_0}\xspace} %

\newcommand{\W}{\mathcal{W}} %
\newcommand{{\QF}}{\mathscr{Q}} %
\newcommand{\refPol}{{\bar{\pol}}} %
\newcommand{\refVal}{\mathcal{V}} %
\newcommand{\temp}{\eta} %
\newcommand{\optRefPol}{{\pol}^*} %

\newcommand{\depth}{\text{depth}} %
\newcommand{\GenModel}{\mathscr{G}} %
\newcommand{\rR}{{{r}}} %
\newcommand{\rV}{{{\mathcal{V}}}} %
\newcommand{\rW}{{{\mathcal{W}}}} %
\newcommand{\rD}{{{\Prob \refVal}}} %

\newcommand{\simd}{\textsc{simd}\xspace}
\newcommand{\simt}{\textsc{simt}\xspace}
\newcommand{\cpu}{\textsc{cpu}\xspace}
\newcommand{\gpu}{\textsc{gpu}\xspace}

\newcommand{\sbmp}{\textsc{sbmp}\xspace}

\newcommand{\sbmps}{\textsc{sbmp}s\xspace}

\newcommand{\rrtc}{\textsc{rrt}-Connect\xspace}

\newcommand{\eg}{\emph{e.g.},\xspace}
\newcommand{\ie}{\emph{i.e.},\xspace}
\newcommand{\dof}{\textsc{d\scalebox{.8}{o}f}\xspace}

\newcommand{\vamp}{\textsc{vamp}\xspace}

\newcommand{\X}{Q}
\newcommand{\Xg}{\X_G}
\newcommand{\Xfree}{\X_{\text{free}}}

\newcommand{\x}{q}

\renewcommand{\xi}{\x_{I}}
\renewcommand{\path}{p}

%% file: sections/introduction.tex
Motion planning in the partially observed and non-deterministic world is a critical component of reliable and robust robot operation. Partially Observable Markov Decision Processes (\pomdps)~\cite{klc1998,SS1973} are a natural way to formulate such problems. The key insight of the \pomdp framework is to represent uncertainty on the effects of actions, perception and initial state as probability distributions, and then reason about the best strategy to perform with respect to distributions over the problem's state space, called \emph{beliefs}, rather than the state space itself. 

Although a \pomdp{}'s systematic reasoning about uncertainty comes at the cost of high computational complexity~\cite{PT1987}, the \pomdp framework is practical for many robotics problems~\cite{kurniawatiSurvey}, thanks in large part to sampling-based approaches. These approaches relax the problem of finding an optimal solution to an approximate one by sampling states from the belief space and computing the best strategies from only the samples. 
Scalable anytime methods under this approach (surveyed in~\cite{kurniawatiSurvey}) have been proposed for solving large \pomdp problems. 
However, computing a good solution to long-horizon (\eg{} $\geq 15$ steps) \pomdps remains difficult. 

Early results from the literature~\cite{KDHL2011} indicate that Sampling-Based Motion Planning (\sbmp)---sampling-based approaches designed for deterministic motion planning---help alleviate the challenges of long-horizon problems in offline \pomdp planning. Specifically, \sbmps can be used to generate suitable macro actions (\ie{} sequences of actions) to reduce the effective planning horizon for a \pomdp solver. Macro actions generated via \sbmp automatically adapt to geometric features of the valid region of the state space and tend to cover a diverse set of macro actions.\looseness=-1 

Although the above approach performs well for offline \pomdp planning, it is often impractical for online planning for two reasons. First, the speed of \sbmp, which historically required hundreds of milliseconds to tens of seconds to find a single motion plan. Second, most online \pomdp planners~\cite{KY13:=ABT,sv10,despot17} exhaustively enumerate each action at each sampled belief in computing the best action to perform. When macro actions are used, exhaustive enumeration is performed over all macro actions; thus, the number of macro actions should be kept low, thereby limiting macro action diversity in online \pomdp planning.\looseness=-1 

However, the recently proposed Vector-Accelerated Motion Planning (\vamp) framework~\cite{tkk23} enables \sbmps to find solutions on microsecond timescales, multiple orders of magnitude faster than prior approaches. 
Concurrently, recently proposed \emph{reference-based \pomdp planners}~\cite{kkk23} remove the requirement for exhaustive enumeration of actions at a small cost to full optimality.
Leveraging these two advances, we propose an online \pomdp solver, called \nopLong (\nop), which is a reference-based \pomdp planner that employs a \vamp-enhanced macro action sampler as its underlying reference policy.

We evaluate \nop on multiple long-horizon \pomdps, including a problem that requires over 100 lookahead steps and a 15-dimensional problem with a planning horizon of over 20. Comparisons with state-of-the-art online \pomdp planners---including \pomcp~\cite{sv10}, a \pomcp modification that uses \vamp to generate macro actions, and \despot~\cite{despot17} with learned macro actions~\cite{lee.rss,KL2023}---indicate \nop substantially outperforms state-of-the-art methods in all evaluation scenarios.

%% file: sections/sbmp.tex
Sampling-based motion planning (\sbmp) is a common, effective family of algorithms (\eg~\cite{Kavraki1996,Kuffner2000}) for solving \emph{deterministic} motion planning problems~\cite{LaValle2006}.
They are empirically able to find collision-free motions for high degree-of-freedom (\dof) robots in environments containing many obstacles by drawing samples from a robot's \emph{configuration space}, the set of all possible robot configurations, $q \in \X$. Constraints, such as (most commonly) avoiding collisions between the robot and the environment or itself, partition the configuration space into \emph{valid}---$\Xfree$---and \emph{invalid}---$\X \setminus \Xfree$---configurations.
A deterministic \emph{motion planning problem} is then a tuple $(\Xfree, \xi, \Xg)$ representing the task of finding a continuous path, $\path: [0, 1] \rightarrow \Xfree$, from an initial configuration, $\xi$, to a goal region, $\Xg \subseteq \Xfree$ (\ie{} $\path(0) = \xi$ and $\path(1) \in \Xg$).

\sbmps{} solve such problems by building a discrete approximation of $\Xfree$ as a graph or tree connecting sampled configurations in $\Xfree$ by short, local motions.
Once both $\xi$ and $\Xg$ are connected by this structure, a valid robot motion plan can be found with graph search methods. The solutions are deterministic, open-loop plans, and may not be robust against uncertainties or changes in configuration spaces.

%% file: sections/pomdp.tex
An infinite-horizon \pomdp is defined as the tuple
$
    \langle \States, \Actions, \Observations, \TP, \OP, \Reward, \discount, \belInit\rangle
$
where $\States$ denotes the set of all possible states of the robot, $\Actions$ denotes the set of all possible robot actions, and $\Observations$ denotes the set of all possible robot observations. 
In our context, the state space encompasses the robot's configuration space. The transition function $\TP(\nst \, | \, \st, \act)$ is the conditional probability that the robot will be in state $\nst \in \States$ after performing action $\act \in \Actions$ from state $\st \in \States$.
The observation function $\OP( \obs \, | \, \nst, \act)$ is the conditional probability that the agent perceives $\obs \in \Observations$ when it is in state $\nst \in \States$ after performing action $\act \in \Actions$.
The reward is a bounded real-valued function $\Reward: \States \times \Actions \rightarrow \Reals$.
The parameter $\discount \in (0,1)$ is the discount factor which ensures the objective function is well-defined. 

In general, the agent does not know the true state. At each time-step, the agent maintains a \emph{belief} about its state---a probability distribution over the state space. The space of all possible beliefs is denoted by $\belSpace:= \simplex(\States)$. The agent starts from a given initial belief, denoted as \belInit. 
If $\belp$ denotes the agent's next belief after taking action $\act \in \Actions$ and receiving the observation $\obs \in \Observations$, then the updated belief is given by
$
\belp(\nst) =\tau(\bel, \act, \obs) \propto \OP(\obs \, | \, \nst, \act) \sum_{\st \in \States} \TP(\nst \, | \, \st, \act) \, \bel(\st)
$.
For any given belief $\bel$ and action $\act$, the expected reward is given by
$\Reward(\bel, \act) := \sum_{\st \in \States} \Reward(\st, \act) \bel(\st)$.

A (stochastic) \emph{policy} is a mapping $\pol: \belSpace \rightarrow \simplex(\Actions)$. We denote its distribution for any given input $\bel \in \belSpace$ by $\pol(\cdot \, | \, \bel)$.
A policy is \emph{deterministic} if it only has support at a single point $\act \in \Actions$.
Let $\policies$ be the class of all policies.
Given a policy $\pol \in \policies$, 
we define the \emph{value function} $\Val^\pol: \belSpace \rightarrow \Reals$ to be the expected total discounted reward, 
$\Val^\pol(\bel) := \Exp [ \sum_{t = 0}^\infty \discount^t \Reward\big(\bel^\pol_t, A^\pol_t)\big) ]$.
In this paper, a \emph{solution} to the \pomdp is a deterministic policy $\optPol \in \policies$ satisfying
$\Val^{\optPol}(\bel) = \sup_{\pol \in \policies} \Val^\pol(\bel)$
on $\belSpace$. 
The Bellman equation
\begin{equation} \label{eq.bellman}
\Val(\bel) = \max_{\act \in \Actions} \Big[ \Reward(\bel, \act) + \discount \sum_{\obs \in \Observations} \Prob(\obs \, | \, \act, \bel) \Val\big( \tau(\bel, \act, \obs) \big)\Big]
\end{equation}
is satisfied by the optimal value function $\Val^{\optPol}$. The intrinsic conditional probability $\Prob( \obs \, | \, \act, \bel)
 := \sum_{\nst \in \States} \OP(\obs \, | \, \nst, \act) ( \sum_{\st \in \States} \TP( \nst \, | \, \st, \act) \bel(\st) )
$
is the probability that the agent perceives $\obs$, having performed the action $\act$, under the belief $\bel$.
Since beliefs are sufficient statistics of the entire history of action-observation pairs from the initial belief $\belInit$~\cite{hauskrecht}, beliefs and histories $h_t:=\{(\act_i, \obs_i)\}_{i=0}^t$ are interchangeable.

%% file: sections/largePOMDPs.tex
Solving long-horizon \pomdps{} remains a significant challenge as the set of possible future events grows exponentially with respect to the planning horizon.
Planning over \emph{macro actions}---a set of action sequences---is a common approach to deal with long-horizon \pomdps{} \cite{FRF2020,he10:puma,KDHL2011,lee.rss,KL2023,theo2003}.
Although this reduces the effective planning horizon, choosing a set of good macro actions for planning is critical and automatic construction of macro actions can require significant computational effort, \eg{} the learning time for \magic~\cite{lee.rss} can be on the order of hours.

Irrespective of whether macro actions are used, most online \pomdp planners~\cite{KY13:=ABT,despot17,sv10} exhaustively enumerate all actions from each sampled belief in order to estimate the optimal action. Most efficient optimization methods, which are generally based on gradient ascent, cannot be used effectively because computing the gradient of the \pomdp{} value function is very expensive. 
Therefore, existing planners seldom explore long-term information and tend to perform poorly for horizons greater than 15 steps.

Recently, the authors of~\cite{kkk23} have mitigated this limitation by solving a \emph{reference-based \pomdp}: a reformulated \pomdp whose reward objective incurs a \kl-penalty for deviating too far from a given stochastic \emph{reference belief-to-belief transition}: $\bar{U}(\act, \obs \mid \bel):=\Prob(\obs \mid \act,\bel)\bar{\pol}(\act \, | \, \bel)$ where $\bar{\pol}$ is a given stochastic reference policy. The reward with respect to $U$ is defined as $\Reward(\bel, U):=\sum_{\act,\obs}\Reward(\bel,\act) \allowbreak U(\act,\obs\mid \bel)$.
The value of a belief is given by
\begin{equation}\label{eq: ref-nips-bellman}
    \rV^*(\bel) = \sup_{U\in\mathcal{U}(b)}\left(\Reward(\bel, U)-\KL(U\,\|\, \bar{U})+\gamma\sum_{a,o}U(a,o\mid b)\rV^*(\tau(b,a,o))\right)
\end{equation}
where $\mathcal{U}(b)\subseteq \Delta(\Actions\times \Observations)$ is the set of admissible transitions at belief $b$.
The \textsc{rhs} can be optimized analytically, effectively removing the need for enumerative optimization in online \pomdp solving and thereby reducing the branching factor of the belief tree.
Preliminary results from~\cite{kkk23} indicate that policies generated by this procedure can outperform state-of-the-art \pomdp benchmarks on long-horizon \pomdps{}.
However, the paper~\cite{kkk23} employs relatively crude reference policies nor does it incorporate macro actions into planning.

%% file: sections/integration-james.tex
We begin by presenting a general mechanism in Algorithm \ref{alg:integration} to integrate belief-space sampling of a canonical online \pomdp planner with state space sampling to generate macro-actions using \sbmp. In principle, Algorithm \ref{alg:integration} can use any \sbmp planner and can generalize to most existing online particle-based \pomdp planners, such as \pomcp \cite{sv10}, \despot \cite{despot17} and their derivatives \cite{hypdespot21,KY13:=ABT,lee.rss,KL2023}.

\begin{algorithm*}[!th]
\caption{Online \pomdp Planner with \sbmp-Generated Macro Actions}
\label{alg:integration}
\begin{algorithmic}[1]
    \State \textbf{Initialize} $T$ from belief particles
    \While {time permitting}
        \State $T \gets \Call{SBMPExpandTree}{T}$
        \State Re-estimate values on $T$ and propagate back up to the root node
    \EndWhile

\end{algorithmic}

\vspace{0.1cm}
\textsc{SBMPExpandTree}($T$)
\begin{algorithmic}[1]
    \State Search $T$ via \sbmp and store node data
    \State Arrive at a (set of) nodes $H$ in $T$ to expand
    \For {$\hist \in H$}
        \State Rapidly sample \sbmp paths $p$ between particles in $\hist$ to selected targets
        \State Add nodes corresponding to actions crafted from $p$ to $T$
        \State Sample new observations and histories, update state particles and add to $T$
        \While {not stopping criteria (e.g. desired depth reached)}
        \State Repeat lines 1--6 on the newly expanded tree $T$
        \EndWhile
    \EndFor
    
\end{algorithmic}

\end{algorithm*}

To infer the best macro action, \textsc{SBMPExpandTree} in Algorithm \ref{alg:integration} samples a set of beliefs reachable from its current belief and represents them as a \emph{belief tree}, $T$---i.e. a tree whose nodes store beliefs (typically represented as sampled state particles) and whose edges represent action-observation pairs.
Methods vary in the construction of the belief tree.
Generally-speaking, online \pomdp planners search the tree forward by judicious sampling---the choice of sampler is the main difference between planners.
When an unvisited node is encountered during the search, existing planners expand every macro action to receive a crude value estimate before propagating newly obtained information back to the root node.

The choice of sampler for forward search is an essential component for achieving high-quality policies online. 
The development of {hardware-accelerated \sbmp} via \simd-vectorization in \cite{tkk23} provides a powerful capability to 
generate computationally cheap, high-quality trajectories in the state space. It enables \sbmp planners to rapidly generate constraint-satisfying (\eg{} collision-free) state space paths to goals or to highly informative states, which 
in turn can be used as macro actions for the \pomdp planner. Thus, promising macro actions can be dynamically created as a subroutine (line 4 in \textsc{SBMPExpandTree}) \emph{within} a \pomdp planner itself in fractions of a second.  

However, fast macro action sampling alone is not sufficient, as current online \pomdp planners perform numerical optimization via enumeration of \emph{all} (macro) actions at each sampled belief, which results in a time and space complexity of  $\bigO(\#\textrm{macro\_actions}^{h})$, where $h$ is the planning horizon. This complexity significantly limits the number of macro actions and
the depth of the tree it can construct, thereby significantly limiting the benefit of \sbmp for \pomdp{} solving.
To tackle this problem, our planner \nop draws on insights from the reference-based \pomdp~\cite{kkk23} formulation while, in tandem, exploiting the speed of \vamp~\cite{tkk23}---an implementation of \simd-vectorized \sbmp---to create a rich set of diverse macro actions, which the planner uses to efficiently explore relevant parts of the belief space.
Section \ref{sec.nop} provides the details of \nop and Section \ref{sec:experiments} empirically verifies the claims made above.

%% file: sections/pop-art.tex
This section introduces our online anytime planner \nopLong (\nop). 
To this end, we motivate it by briefly outlining \vamp's capability to induce high-quality reference policies before adapting the reference-based \pomdp formulation from \cite{kkk23} to our context. The algorithm is then constructed on top of these components.

\subsection{Vector Accelerated Motion Planning (VAMP)}
\label{subsec.vamp}
\input{sections/vamp}

\subsection{Reference-Based POMDPs over Stochastic Actions} \label{subsec.rbpomdp}
\input{sections/refPomdp}

\subsection{\nopTitle}

\input{sections/algorithm}

%% file: sections/vamp.tex
Towards achieving fast motion planning, recent works have introduced new perspectives on \emph{hardware-accelerated} \sbmps, using either \cpu{} single-instruction, multiple-data (\simd)~\cite{tkk23} or \gpu{} single-instruction, multiple-thread (\simt)~\cite{sundaralingam2023curobo} parallelism to find complete motion plans in tens of microseconds to tens of milliseconds.
In particular, the authors of \vamp \cite{tkk23} proposed a \simd-vectorized approach to computing \sbmp{} primitives (\ie{} local motion validation) that applies to all \sbmps{} and is available on any modern computer without specialized hardware; thus, it is now possible to generate probabilistically-complete, global, collision-free trajectories for high-\dof systems at kilohertz rates---on the scale of tens of thousands of plans per second.
The key insight of this work is to lift the ``primitive'' operations of the \sbmp to operate over \emph{vectors} of configurations in parallel.
Functionally, this enables checking validity of a spatially distributed set of configurations over a candidate motion in parallel for the cost of a single collision check, massively lowering the expected time it takes to find colliding configurations along said motion. This development has called into question previously held perceptions that \sbmps{} are relatively time-expensive subroutines and---in the context of planning under uncertainty---opens the door to using \sbmps{} to guide \pomdp planning on the fly.

%% file: sections/refPomdp.tex
The concept of a Reference-Based \pomdp, as reviewed in Sect. \ref{sec: long horizon POMDPs}, was introduced in \cite{kkk23} as a generalization of the \mdp formulations using \kl-penalization in~\cite{azar12,todorov} to \pomdp{}s.
One limitation is that the formulation given by \eqref{eq: ref-nips-bellman} is somewhat artificial in that reference policies are stated with respect to \emph{belief-to-belief} transitions (see discussion in Sect. 8 of~\cite{kkk23}).
In this paper, we will use a more natural formulation of a Reference-Based \pomdp over \emph{stochastic actions} rather than \emph{belief-to-belief transitions} as in~\cite{kkk23}.
Note that in doing so, all the benefits of the formulation in~\cite{kkk23} (as detailed in Sect. \ref{sec: long horizon POMDPs}) are still preserved.

Specifically, a Reference-Based \pomdp over stochastic actions is completely specified by the tuple
$\langle \States, \Actions, \Observations, \OP, \TP, \Reward, \discount, \temp, \refPol \rangle$ where, in addition to the standard data, we have a \emph{temperature} parameter $\temp > 0$ and a given (stochastic) \emph{reference policy} 
$\refPol(\cdot \, | \, \bel)$.
Its value
$\refVal$ for a given $\bel \in \belSpace$ satisfies the Bellman equation
\begin{equation} \label{eq.ref.bellman}
        \refVal(\bel) = \sup_{\pol \in \policies} \Big[ \Reward(\bel, \pol) - \frac{1}{\temp} \KL( \pol \, \| \, \refPol )
        + \discount \sum_{\act, \obs} \Prob(\obs \, | \, \act, \bel) \pol(\act \, | \, \bel) \refVal\big( \tau( \bel, \act, \obs) \big) \Big]
\end{equation}
where $\Reward(\bel, \pol) := \sum_{\act, \st} \Reward(\st, \act) \pol(\act \, | \, \bel) \bel(\st) $ is the reward estimate.
A \emph{solution} is a stochastic policy $\pol \in \policies$ that maximizes the value.
The problem can therefore be viewed as a \kl-penalized \pomdp whose objective is modified to trade off two (potentially competing) objectives: (1) abide by the reference policy, and (2) maximize reward.
The trade-off is balanced by $\temp$ and the \emph{quality} of the reference policy---higher-quality reference policies are those that reduce the \kl-divergence between the solution of the unpenalized \pomdp \eqref{eq.bellman} and the reference policy.
The supremum in \eqref{eq.ref.bellman} can be attained analytically by extending an argument of~\cite{azar11,azar12} to \pomdps.
This yields an equivalent form of \eqref{eq.ref.bellman}---\ie{}
\begin{equation} \label{eq.maximized}
    \refVal(\bel) = \frac{1}{\temp} \log \Big[\sum_\act \refPol(\act \, | \, \bel) \exp \Big\{ \temp \big[ \Reward(\bel, \act) + \discount \sum_\obs \Prob(\obs \, | \, \act, \bel) \refVal\big( \tau(\bel, \act, \obs) \big) \big] \Big\} \Big].
\end{equation}
Moreover, the \emph{exact} solution of the Reference-Based \pomdp is given by
\begin{equation} \label{eq.rbpomdp.opt.pol}
\optRefPol(\act \, | \, \bel) 
\propto {\refPol(\act \, | \, \bel) \exp\Big\{\temp  \Big[ \Reward(\bel, \act) + \discount \sum_\obs \Prob(\obs \, | \, \act, \bel) \refVal^*\big(\tau(\bel, \act, \obs)\big) \Big] \Big\}}.
\end{equation}
The main point is that enumerative maximization can be avoided; instead, the solution can be approximated by successively iterating the analytic Bellman backup \eqref{eq.maximized} exactly. This procedure is guaranteed to converge to a unique solution $\refVal^*$ from which the policy can be read off from \eqref{eq.rbpomdp.opt.pol}.
Unfortunately, needing \emph{exact} backups is a significant requirement, given the cost of computing exact beliefs and, consequently, the immediate expected rewards $\Reward(\bel, \act)$ for every node in the belief tree.
Still, the expectations 
$
    \rD(\bel, \act) := \sum_\obs \Prob(\obs \, | \, \act, \bel) \refVal\big( \tau (\bel, \act, \obs) \big)
$
and
$
     \W(\bel) := \sum_\act \refPol(\act \, | \, \bel) \exp \{ \temp [ \Reward(\bel, \act) + \discount \rD(\bel, \act) ] \}
$
can be approximated by \emph{sampling} the reference policy and generative model.
Improving value estimates is extremely efficient as backups can be computed by \emph{maintaining} sums rather than enumeratively maximizing over actions.

Even so, the simplicity offered by the formulation comes at a cost. If the optimal policy of the \pomdp with an unmodified objective is too far from the reference policy (in the sense of the \kl-divergence), pure reward maximization can be compromised.
Of course, the optimal policy and hence this divergence are not known a priori. 
Instead \nop assumes that reference policies generated by accelerated \sbmps provide a reasonable starting point (see Sect \ref{subsec.vamp}), leveraging them to rapidly sample high-quality deterministic policies to induce a reasonable partially observed reference policy which it deforms online.
We emphasize that this modification is not negligible.
Indeed, our results in Sect. \ref{sec:experiments} show that the performance deteriorates with problem complexity if the agent only executes the \sbmp-induced reference policy as an open-controller without accounting for uncertainty.

%% file: sections/algorithm.tex
\nop (Algorithm \ref{alg: ref-prm}) adds new history nodes $\hist$ to the search tree by sampling macro actions $\mact$ from a reference policy induced from a \simd-vectorized \sbmp up to a predefined depth $D$, keeping track of the states $\st$, (macro) observations $\mobs$ and rewards $r$ sampled under a \emph{generative model} $\GenModel$---\ie{} a simulator for the \pomdp environment.
When the required depth is reached, \nop obtains a crude estimate of the root node's value via a value heuristic. The planner then approximates the exact backup (\textsc{MaintainExpectation}) by carefully \emph{maintaining} an empirical expectation, repeating backups on the simulated belief-tree path up to the root node---this form of backup is justified by our discussion in Sect. \ref{sec.nop} and, in particular, Eq. \eqref{eq.maximized}.
The above procedure is repeated until timeout, at which point the optimal policy's empirical estimate is read off from the tree's root node and executed---the form of estimate is given by \eqref{eq.rbpomdp.opt.pol}.

\begin{algorithm*}[h]
\caption{\textsc{\nop}}
\label{alg: ref-prm}
\begin{algorithmic}[1]
\State Initialize tree $T$ rooted at $\hist$
\While {time permitting}
    \State \Call{Simulate}{$\hist$}
\EndWhile
\State \Return $T$
\end{algorithmic}

\vspace{0.1cm}
\textsc{Simulate}($\hist$)
\begin{algorithmic}[1]
\State Sample $\st$ from belief particles of $\hist$
\If {$\depth(\hist) > D$}
    \State \Return \Call{ValueHeuristic}{$\hist, \st$}
\EndIf
\State $\mact \gets \Call{SampleMacroActionSBMP}{\hist, \st}$
\State Sample $(\nst, \mobs, r(\st, \mact; \discount))$ from generative model $\GenModel(\st, \mact)$
\State Create nodes for $\hist \mact$ and $\hist \mact \mobs$ if not created already
\State Add $\nst$ to belief particles of $\hist \mact \mobs$
\State \Return $\rV(\hist) \gets \log(\Call{MaintainExpectation}{\hist, \mact, \mobs, r}) / \temp$
\end{algorithmic}

\vspace{0.1cm}
\textsc{SampleMacroActionSBMP}($\hist, \st$)
\begin{algorithmic}[1]
    \State Get the current configuration $q_{\text{start}}$ from $\st$ (assert free)
    \State  $q_{\text{goal}} \gets \samplingHeuristic()$ (assert free)
    \State Plan path $p$ from $q_{\text{start}}$ to $q_{\text{goal}}$ using fast \sbmp
    \State \Return (macro) action $\mact$ ``fashioned'' from $p$
\end{algorithmic}

\vspace{0.1cm}
\textsc{MaintainExpectation}($\hist, \mact, \mobs, r$)
\begin{algorithmic}[1]
    \State $w \gets N(\hist)  \rW(\hist) - N(\hist \mact) \exp\left(\temp\left[\rR(\hist \mact) + \discount^{|\mact|} \rD(\hist \mact)\right]\right)$
    \State $p \gets N(\hist \mact) \rD(\hist \mact) - N(\hist \mact \mobs) \rV(\hist \mact \mobs)$
    \State $N(\hist \mact) \gets N(\hist \mact) + 1$
    \State $\rR(\hist \mact) \gets \rR(\hist \mact) + {r - \rR(\hist \mact)}/{N(\hist \mact)}$
    \State $\rD(\hist \mact) \gets \big( p +  N(\hist \mact \mobs) \, \Call{Simulate}{\hist \mact \mobs} \big)/ {N(\hist \mact)}$
    \State $N(\hist) \gets N(\hist) + 1$
    \State \Return $\rW(\hist) \gets \Big\{ w + N(\hist \mact) \exp\left(\temp\left[ \rR(\hist \mact) + \discount^{|\mact|} \rD(\hist \mact)\right]\right)\Big\}/{N(\hist)}$
\end{algorithmic}

\end{algorithm*}

The subroutine \textsc{SampleMacroActionSBMP} leverages \vamp as the reference policy to sample macro actions. Our implementation of \nop instantiates this subroutine with an instance of \samplingHeuristic which selects meaningful target points, towards which \vamp's planners can be used to rapidly generate a path.
For environments with well-defined goal and informative states, examples of \samplingHeuristic include:
\begin{myItemize}
    \item \textbf{Uniform} Uniformly sample goal configurations with a given probability; otherwise, sample an informative state configuration uniformly.
    \item \textbf{Distance}. Uniformly sample a goal configuration with a given probability; otherwise, sample an informative state with a probability inversely proportional to a distance between the configurations of the current and informative states.
    \item \textbf{Entropy}. Given the normalized belief entropy $\mathcal{H}$, sample goal configurations with probability $1-\mathcal{H}$; otherwise sample remaining informative configurations with probability inverse to a distance between the configurations of the current and informative states.
\end{myItemize}

%% file: sections/experiments.tex
We evaluated \nop on four different long-horizon \pomdp problems systematically analyzing the effects of its respective components on its performance.

\subsection{Scenarios and Benchmark Methods}

\begin{figure}[t]
        \centering
        \begin{subfigure}[b]{0.35\textwidth}
            \centering
            \includegraphics[width=\textwidth]{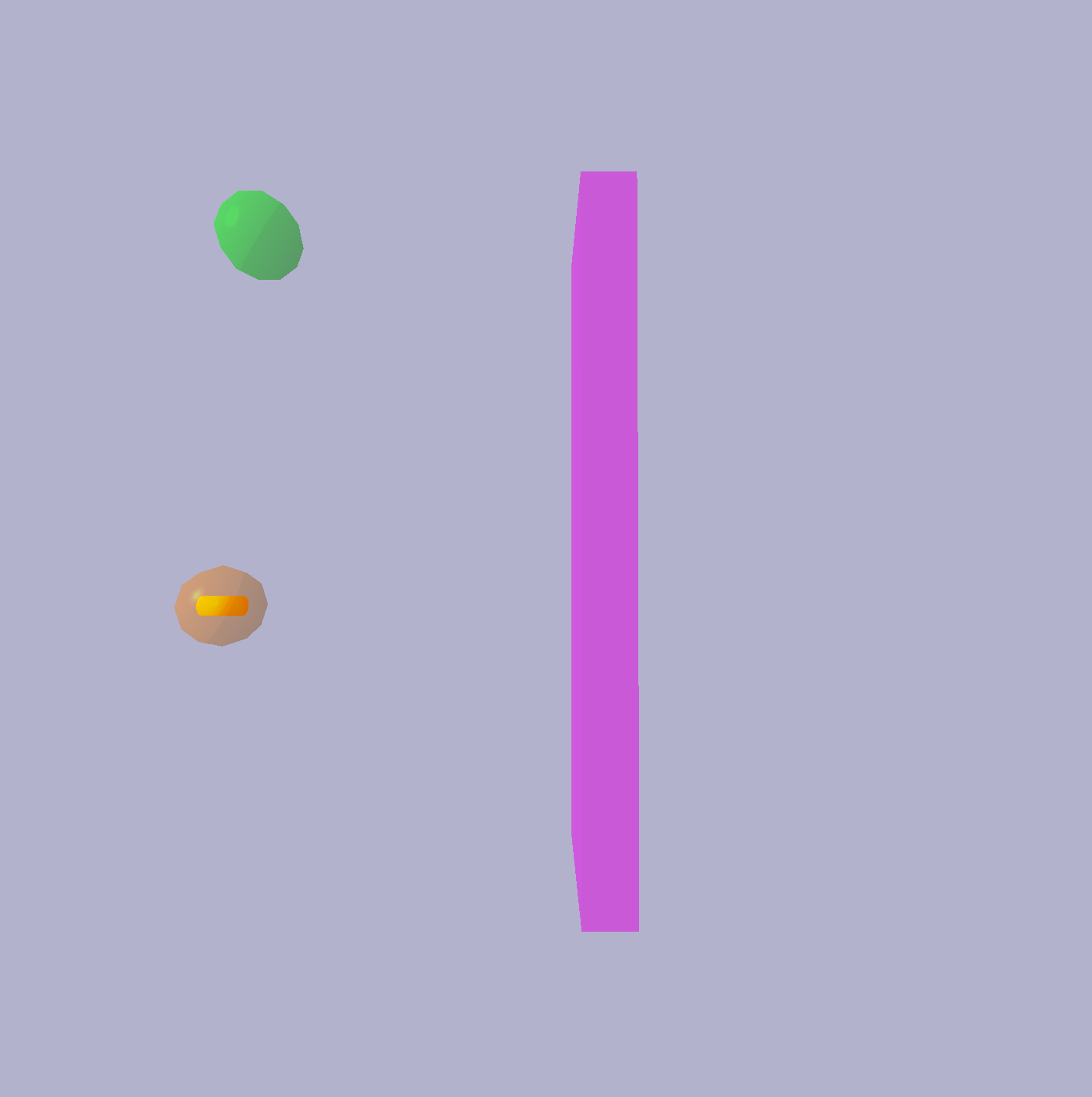}
            \caption[Light-Dark]%
            {{\small Light-Dark}}    
            \label{fig: visual light dark}
        \end{subfigure}
        \begin{subfigure}[b]{0.35\textwidth}  
            \centering 
            \includegraphics[width=\textwidth]{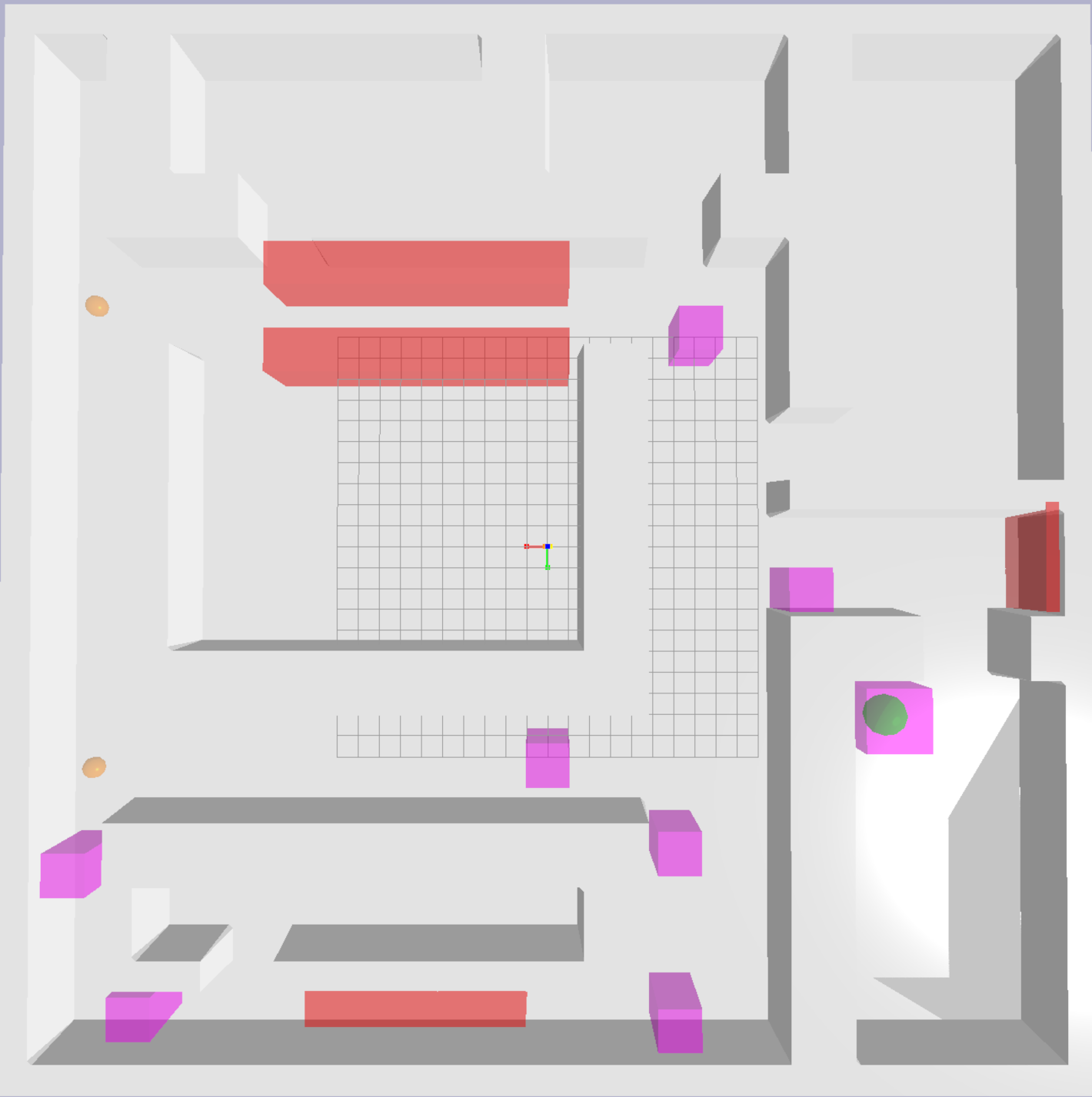}
            \caption[]%
            {{\small Maze2D}}    
            \label{fig: visual maze 2D}
        \end{subfigure}
        \begin{subfigure}[b]{0.35\textwidth}   
           \centering 
            \includegraphics[width=\textwidth]{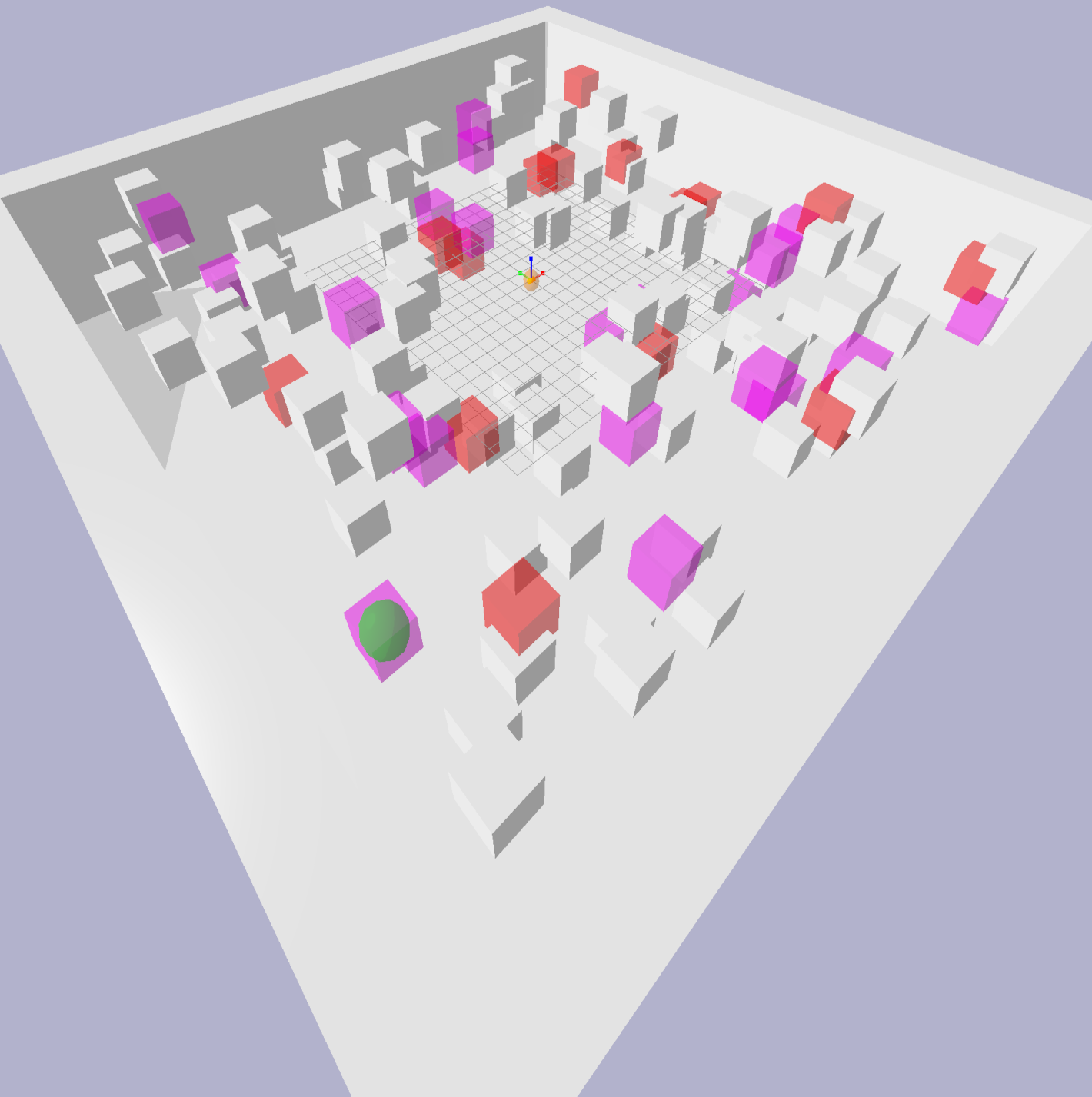}
            \caption[]%
            {{\small Random3D}}    
            \label{fig:visual Random 3D}
        \end{subfigure}
        \begin{subfigure}[b]{0.35\textwidth}   
            \centering 
            \includegraphics[width=\textwidth]{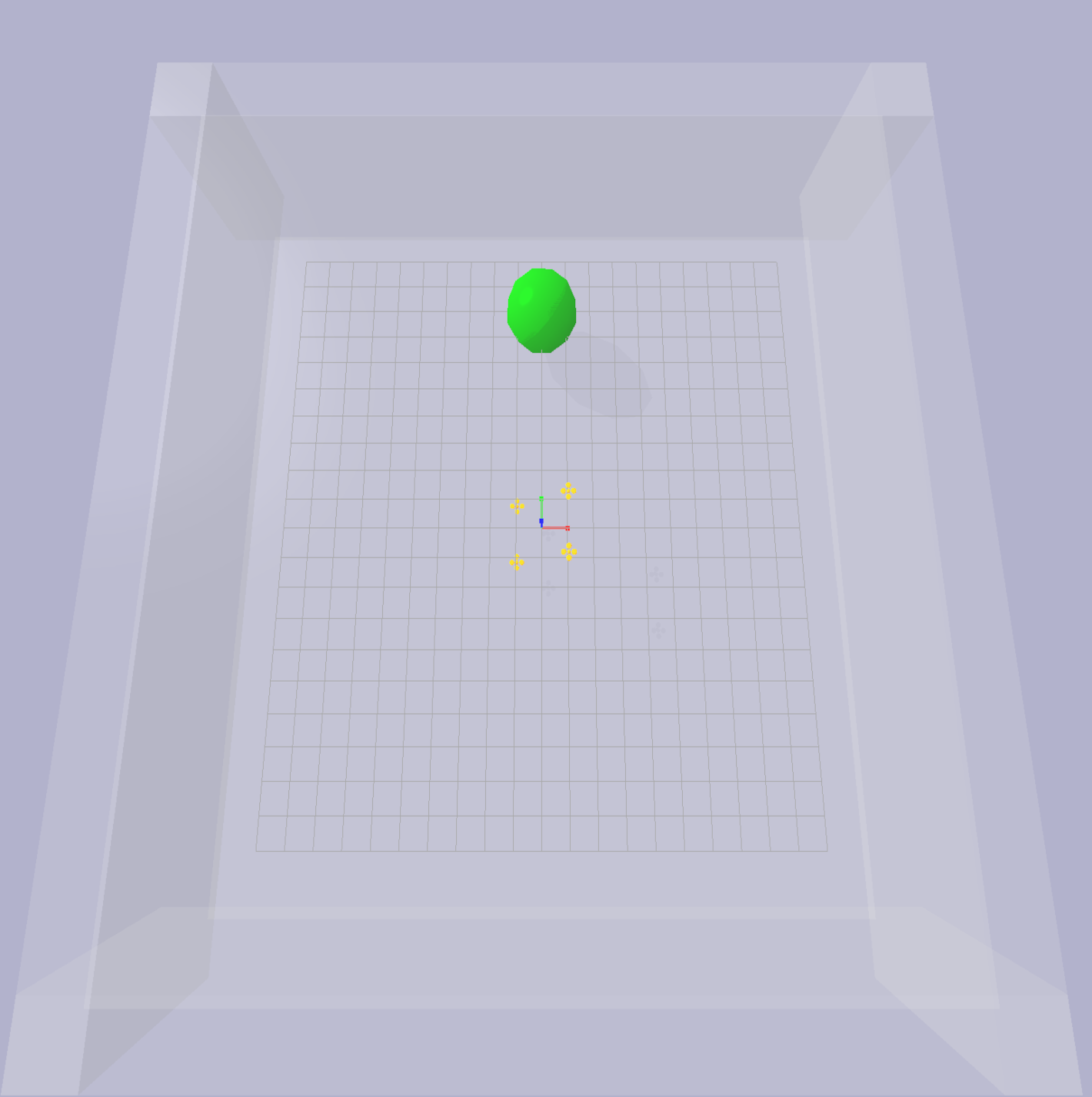}
            \caption[]%
            {{\footnotesize Multi-Drone Tag}}    
            \label{fig:visual capture}
        \end{subfigure}
        \caption[]
        {\small Benchmark environments with obstacles (\textcolor{gray}{\textbf{grey}}), landmarks (\textcolor{purple}{\textbf{purple}}), danger zones (\textcolor{red}{\textbf{red}}), starting locations (\textcolor{orange}{\textbf{orange}}) and goals and targets (\textcolor{green}{\textbf{green}})
        }
        \label{fig: Visualisations}
\end{figure}

\begin{myDescription}
	\item[Light-Dark (\fref{fig: visual light dark}).]~ A variation of the classical Light-Dark problem. The state and observation space are continuous and identical to those in~\cite{lee.rss,KL2023,platt2010belief}, but we modify the problem to make the action space discrete---at each step, the robot can move in four cardinal directions with a fixed step-size of 0.5. Moreover, the agent receives feedback when it reaches the goal. The step and goal rewards are -0.1 and 100 respectively. The light stripe, the goal and the initial positions are randomly drawn from an $8 \times 8$ unit box but are constrained to be 4 units away from each other. Hence, a minimum of 8 primitive actions are required to navigate to the goal.
	\item[Maze2D (\fref{fig: visual maze 2D}).]~This is a very long-horizon problem, modified from the discrete 2D Navigation scenario in \cite{KDHL2011}. A 2-\dof mobile cuboid robot needs to navigate from one of the spawn points to a goal region without colliding with obstacles or entering a danger zone. The robot's state is its position on the map ($\States = [0, 50]^2$). At each time step, the robot can move in one of the four cardinal directions ($|\Actions|=4$) with a fixed step size, but there is a 20\% chance that the robot moves in either direction orthogonal to the intended direction; if the resulting action would cause a collision with an obstacle, it does not move. The robot can only localize itself in landmark regions receiving position readings with small Gaussian noise inside the landmarks and no observations otherwise. The two spawn points are marked by orange circles in the maze. Each step costs the robot -0.1, the danger zone penalty is -800 and the goal reward is 800. To perform well over the long horizon---a minimum of 100 primitive actions is required to reach the goal---this scenario requires the robot to take detours to localize and avoid danger zones before navigating to the goal.
	\item[Random3D (\fref{fig:visual Random 3D}).]~This is a 3D navigation problem with uniformly random obstacles, landmarks, danger zones and goals; the only constraint is that the goal needs to be at least 40 units from the robot's spawn position at the map's center and a valid path must exist to the goal.
 The \pomdp{}'s state and action spaces are respectively  $\States = [0, 50]^2 \times [0, 6] \subset \mathbb{R}^3$ and the 3D cardinal directions $\mathcal{A}=\{\text{North}, \text{South}, \text{East}, \text{West}, \text{Up}, \text{Down}\}$. The observation, transition and reward models are essentially the same as that of Maze2D except that the error directions are randomized from the 3D cardinal ones.
 Our aim is to systematically evaluate \nop on environments with progressively increasing obstacle density.
     The chances that the \sbmp fails to find a path within a given time limit increases with the obstacle density and a \pomdp planner needs to consider more diverse macro actions to find a good motion strategy.   
	\item[Multi-Drone with Teleporting Target (\fref{fig:visual capture}).]~This scenario with a high-dimensional state space and complex environment dynamics is similar to the multi-robot tag and predator-prey problems considered in~\cite{ong2010planning,schwartz2024posggym}. Four drones, initialized at the center of the environment, need to work together to firstly detect and capture a point target (in green) whose initial position is not known to the drones. 
 The \pomdp{}'s state space is $\States = ([0, 30]^2 \times [0, 4] )^5 \subset \mathbb{R}^{15}$ consisting of the valid positions of the holonomic drones and an evading target. The action space is the space of all 4-tuples consisting of the 3D cardinal directions ($|\Actions| = 24$).
 Both the drones and the target move with a step size of 0.5.
 The target has a detection radius of 4; if any drones are within this range, the target will move in the direction furthest from the closest drone's location. Otherwise, it moves randomly. Each drone has a detection radius of 5 units and receives noisy readings of the target's position when the target is within this radius, otherwise no observation is received. If any drone receives an observation, all drones share this collective knowledge. The target is captured when at least one drone is within 1.5 units distance from the target. However, the target can \textit{teleport} to the opposite of the map once it collides with the map boundaries but drones cannot. This problem requires drones to cooperate to capture the target by either surrounding it within the map's boundaries or by anticipating teleportation elsewhere. A reward of 500 is given if the target is captured, otherwise a -0.1 penalty is incurred by the drones for each step. A run terminates after 40 steps. This is a long-horizon problem---even if the target's initial position is known, at least 20 primitive steps are required to capture the tag.
\end{myDescription}

We compare \nop with several instantiations of Algorithm \ref{alg:integration} with state-of-the-art \pomdp solvers, as well as our reference-based instantiation \nop. Specifically, we compare with the following:
\begin{myDescription}
    \item[Belief-VAMP (B-VAMP)]. The planner maintains the belief of the current state of the agent but only takes actions returned by \vamp{}'s macro action sampler \emph{without} planning.
    \item[Ref-Basic]. The reference-based \pomdp planner from~\cite{kkk23} (no \vamp).
    \item[POMCP~\cite{sv10}]. A standard benchmark for online \pomdp planning.
    \item[R-POMCP]. An instantiation of Algorithm \ref{alg:integration} using \pomcp with macro actions generated by \vamp in the same way as \nop, but a finite set of macro actions are fixed for each belief node over which \pomcp optimizes.
    \item[MAGIC~\cite{lee.rss}]. A variation of \despot~\cite{despot17} that uses learnable macro actions generated via an actor-critic approach. \magic uses fixed-length macro actions, so comparisons are for varying lengths (e.g. \magic-x means the macro action length is x).
    \item[RMAG~\cite{KL2023}]. \despot with macro action learning boosted by recurrent neural networks. We compare against \rmag over different macro-actions lengths.
\end{myDescription}

\subsection{Experimental Setup}
For evaluation, all variants of \nop are implemented in Cython following~\cite{zheng2020pomdp_py}. \vamp is implemented in C++ and is used via a Python \textsc{api}. We avoid implementing benchmark methods from scratch for a fairer comparison; the \pomcp implementation is from~\cite{zheng2020pomdp_py} and the implementations of \pomcp, \magic, \rmag and Ref-Basic all use the code implemented by their respective authors~\cite{lee.rss,KL2023}.

All methods are provided with the same planning time of \textbf{1s} across scenarios with the exception of Light-Dark where the planning time is \textbf{0.1s}. We also use the same discount factor ($\gamma=0.99$) for all planners. The temperature parameter $\eta$ for \nop and Ref-Basic is $\eta=0.2$. \magic and \rmag are trained on a 4070 GPU with data collected across 500,000 runs ($\sim$3 hours of training).  Parameters have been tuned to optimize performance for each problem. Aforementioned sampling heuristics (see Section 4) are tested in all environments where  appropriate. For the multi-drone environment, the sampling heuristic randomly samples a state from the belief. For that sampled state, the drone nearest to the target converges to it while other drones spread to a uniformly sampled free configuration.

\subsection{Results and Discussions}

We ran the method 30$\times$ for each scenario and method. The results are summarized in Tables~\ref{tab:2dexperimentsresults},~\ref{tab:maze3dresults}, and~\ref{tab:dronecaptureresults}. Note that the number of episodes and steps recorded in the tables are average statistics of the methods regardless of success or failure. \magic and \rmag were only run on Light-Dark and Maze2D because they do not naturally extend to high-dimensional state spaces.
\pomcp and Ref-Basic failed on all runs of Random3D and Multi Drone Tag and are therefore excluded from the tables.

\begin{table}[!htbp]
\caption{Results on Light-Dark and Maze2D}
\label{tab:2dexperimentsresults}
{\small (Light-Dark: x=4, y=8, z=16. Maze2D: x=8, y=16, z=24)}
\centering
\resizebox{\textwidth}{!}{
\begin{tabular}{cc|cccc|cccc|}
\cline{3-10}
                                          &                     & \multicolumn{4}{c|}{\textbf{Light-Dark}}                                                                                                                                                                                                    & \multicolumn{4}{c|}{\textbf{Maze2D}}                                                                                                                                                                                                         \\ \hline
\multicolumn{1}{|c|}{}   & \textbf{\begin{tabular}[c]{@{}c@{}}Sampling\\ Heuristics\end{tabular}} & \multicolumn{1}{c|}{\textbf{\begin{tabular}[c]{@{}c@{}}\#\\ Episodes\end{tabular}}} & \multicolumn{1}{c|}{\textbf{\begin{tabular}[c]{@{}c@{}}Succ.\\ \%\end{tabular}}} & \multicolumn{1}{c|}{\textbf{\begin{tabular}[c]{@{}c@{}}E[Tot.Reward]\\ (stdErr)\end{tabular}}} & \textbf{Steps}            & \multicolumn{1}{c|}{\textbf{\begin{tabular}[c]{@{}c@{}}\#\\ Episodes\end{tabular}}} & \multicolumn{1}{c|}{\textbf{\begin{tabular}[c]{@{}c@{}}Succ.\\ \%\end{tabular}}} & \multicolumn{1}{c|}{\textbf{\begin{tabular}[c]{@{}c@{}}E[Tot.Reward]\\(stdErr)\end{tabular}}} & \textbf{Steps}             \\ \hline
\multicolumn{1}{|c|}{B-\vamp}                 & entropy             & \multicolumn{1}{c|}{N/A}               & \multicolumn{1}{c|}{76}                                                          & \multicolumn{1}{c|}{70.2 (7.9)}                                                     & 46                        & \multicolumn{1}{c|}{N/A}               & \multicolumn{1}{c|}{40}                                                          & \multicolumn{1}{c|}{-65.6}                                                          & 385                        \\ \hline
\multicolumn{1}{|c|}{\pomcp}               & N/A                 & \multicolumn{1}{c|}{282}               & \multicolumn{1}{c|}{77}                                                          & \multicolumn{1}{c|}{72.6 (7.9)}                                                     & 42                        & \multicolumn{1}{c|}{314}               & \multicolumn{1}{c|}{0}                                                           & \multicolumn{1}{c|}{-80 (0)}                                                        & 800                        \\ \hline
\multicolumn{1}{|c|}{Ref-Basic}           & entropy             & \multicolumn{1}{c|}{44}                & \multicolumn{1}{c|}{50}                                                          & \multicolumn{1}{c|}{45.1 (9.3)}                                                     & 49                        & \multicolumn{1}{c|}{33}                & \multicolumn{1}{c|}{0}                                                           & \multicolumn{1}{c|}{-807 (3.2)}                                                     & {\color[HTML]{FE0000} 75}  \\ \hline
\multicolumn{1}{|c|}{R-\pomcp}               & N/A                 & \multicolumn{1}{c|}{262}               & \multicolumn{1}{c|}{83}                                                          & \multicolumn{1}{c|}{79 (6.9)}                                                     & 31                        & \multicolumn{1}{c|}{279}               & \multicolumn{1}{c|}{0}                                                           & \multicolumn{1}{c|}{-146 (0)}                                                        & 784                        \\ \hline
\multicolumn{1}{|c|}{\magic-x}             & N/A                 & \multicolumn{1}{c|}{N/A}               & \multicolumn{1}{c|}{88}                                                          & \multicolumn{1}{c|}{85.0 (1.1)}                                                     & 29                        & \multicolumn{1}{c|}{N/A}               & \multicolumn{1}{c|}{0}                                                           & \multicolumn{1}{c|}{-80 (11)}                                                       & 800                        \\ \hline
\multicolumn{1}{|c|}{\magic-y}             & N/A                 & \multicolumn{1}{c|}{N/A}               & \multicolumn{1}{c|}{85}                                                          & \multicolumn{1}{c|}{81.9 (1.2)}                                                     & 31                        & \multicolumn{1}{c|}{N/A}               & \multicolumn{1}{c|}{0}                                                           & \multicolumn{1}{c|}{-106 (17)}                                                      & 800                        \\ \hline
\multicolumn{1}{|c|}{\magic-z}             & N/A                 & \multicolumn{1}{c|}{N/A}               & \multicolumn{1}{c|}{78}                                                          & \multicolumn{1}{c|}{75.9 (1.3)}                                                     & 35                        & \multicolumn{1}{c|}{N/A}               & \multicolumn{1}{c|}{1.1}                                                         & \multicolumn{1}{c|}{-408 (12)}                                                      & 716                        \\ \hline
\multicolumn{1}{|c|}{\rmag-x}              & N/A                 & \multicolumn{1}{c|}{N/A}               & \multicolumn{1}{c|}{88}                                                          & \multicolumn{1}{c|}{84.8 (1.1)}                                                     & 29                        & \multicolumn{1}{c|}{N/A}               & \multicolumn{1}{c|}{0}                                                           & \multicolumn{1}{c|}{-80 (0)}                                                        & 800                        \\ \hline
\multicolumn{1}{|c|}{\rmag-y}              & N/A                 & \multicolumn{1}{c|}{N/A}               & \multicolumn{1}{c|}{83}                                                          & \multicolumn{1}{c|}{80.4 (1.2)}                                                     & 32                        & \multicolumn{1}{c|}{N/A}               & \multicolumn{1}{c|}{7.2}                                                         & \multicolumn{1}{c|}{244 (13)}                                                       & 586                        \\ \hline
\multicolumn{1}{|c|}{\rmag-z}              & N/A                 & \multicolumn{1}{c|}{N/A}               & \multicolumn{1}{c|}{80}                                                          & \multicolumn{1}{c|}{77.1 (1.3)}                                                     & 35                        & \multicolumn{1}{c|}{N/A}               & \multicolumn{1}{c|}{0}                                                           & \multicolumn{1}{c|}{-80 (0)}                                                        & 800                        \\ \hline
\multicolumn{1}{|c|}{\nop} & uniform             & \multicolumn{1}{c|}{124}               & \multicolumn{1}{c|}{{\color[HTML]{FE0000} 100}}                                  & \multicolumn{1}{c|}{{\color[HTML]{FE0000} 97.2 (0.1)}}                              & {\color[HTML]{FE0000} 20} & \multicolumn{1}{c|}{115}               & \multicolumn{1}{c|}{ 90}                                   & \multicolumn{1}{c|}{ 596 (57)}                                & 319                        \\ \hline
\multicolumn{1}{|c|}{\nop} & distance            & \multicolumn{1}{c|}{119}               & \multicolumn{1}{c|}{{\color[HTML]{3531FF} 97}}                                   & \multicolumn{1}{c|}{94.2 (3.3)}                                                     & 24                        & \multicolumn{1}{c|}{253}               & \multicolumn{1}{c|}{{\color[HTML]{3531FF} 93}}                                   & \multicolumn{1}{c|}{{\color[HTML]{3531FF}698 (38)}}                                                       & 382                        \\ \hline
\multicolumn{1}{|c|}{\nop} & entropy             & \multicolumn{1}{c|}{26}                & \multicolumn{1}{c|}{{\color[HTML]{FE0000} 100}}                                  & \multicolumn{1}{c|}{{\color[HTML]{3531FF} 97.0 (0.3)}}                              & {\color[HTML]{3531FF} 23} & \multicolumn{1}{c|}{46}                & \multicolumn{1}{c|}{{\color[HTML]{FE0000} 100}}                                   & \multicolumn{1}{c|}{{\color[HTML]{FE0000} 761 (2.9)}}                                & {\color[HTML]{3531FF} 293} \\ \hline
\end{tabular}}
\end{table}

The results indicate that all variants of \nop substantially outperform all baseline methods in all evaluation scenarios. The improvement provided by \nop is smallest for Light-Dark (the simplest evaluation scenario) where all variants of \nop achieved a success rate of up to 28\% higher than \rpomcp, \magic, and \rmag.
In the other scenarios, all variants of \nop improved the success rate of the benchmark methods by many folds. The reason is Light-Dark requires much lower planning horizon and has less uncertainty, compared to other scenarios. This simplicity is indicated by the good performance of methods that do not use macro actions, such as \pomcp and Ref-Basic, and by the good performance of B-\vamp, which reasons with respect to only a sampled state of the belief. 

However, when the scenario requires a much longer planning horizon (e.g., Maze2D) or has higher uncertainty and more complex geometric structures, the benefit of \nop increases. By using \vamp, \nop can quickly generate much more diverse macro actions that capture the geometric features of the problem well. Simultaneously, the reference-based \pomdp solver enables \nop to utilize the sampled macro-actions more efficiently than other benchmark methods. Learning-based methods perform poorly in Maze2D due to the difficulty in learning suitable macro actions with fixed length. In contrast, by using \sbmp, \nop is able to better capture the twists and turns the robot needs to perform to navigate the geometric features of the environment.

\begin{table}[!htbp]
\caption{Results on Random3D}
{\footnotesize(\#1: 100 obstacles, 15 danger zones. \#2: 200 obstacles, 10 danger zones. \\ \#3: 300 obstacles, 10 danger zones. \#4: 400 obstacles, 5 danger zones.)}
\label{tab:maze3dresults}
\centering
\resizebox{\textwidth}{!}{
\begin{tabular}{|c|c|c|c|c|c|c|c|c|c|c|c|}
\hline
        & \textbf{\begin{tabular}[c]{@{}c@{}}Sampling\\ Heuristics\end{tabular}} & \textbf{\begin{tabular}[c]{@{}c@{}}Exp\\ \#\end{tabular}} & \textbf{\begin{tabular}[c]{@{}c@{}}Ref.\\ Policy\\ Fail \%\end{tabular}} & \textbf{\begin{tabular}[c]{@{}c@{}}Success\\ \%\end{tabular}} & \textbf{\begin{tabular}[c]{@{}c@{}}E[Tot.Reward]\\(stdErr)\end{tabular}} & \textbf{\begin{tabular}[c]{@{}c@{}}Steps\end{tabular}} & \textbf{\begin{tabular}[c]{@{}c@{}}Exp\\ \#\end{tabular}} & \textbf{\begin{tabular}[c]{@{}c@{}}Ref.\\ Policy\\ Fail \%\end{tabular}} & \textbf{\begin{tabular}[c]{@{}c@{}}Success\\ \%\end{tabular}} & \textbf{\begin{tabular}[c]{@{}c@{}}E[Tot. Reward]\\(stdErr)\end{tabular}} & \textbf{\begin{tabular}[c]{@{}c@{}}Steps\end{tabular}} \\ \hline
B-\vamp     & N/A                 &   \multirow{5}{*}{1}                                                             & N/A                                                                      & 0                                                             & -632 (52)                                                     & 435                                                            &        \multirow{5}{*}{2}                                                        & N/A                                                                      & 10                                                            & -307 (65)                                                     & 526                                                            \\ \cline{1-2} \cline{4-7} \cline{9-12}
R-\pomcp     & N/A                 &                                                                & N/A                                                                      & 7                                                             & -89 (52.2)                                                     & 741                                                            &                                                                & N/A                                                                      & 7                                                            & -157 (64)                                                     & 700                                                            \\ \cline{1-2} \cline{4-7} \cline{9-12}
\nop & uniform             &                                      & 6                                                                        & \textcolor{blue}{60}                                                            & \textcolor{blue}{236} (121)                                                     & \textcolor{blue}{287}                                                            &                                      & 12                                                                       & \textcolor{red}{63}                                                            & \textcolor{red}{224} (123)                                                     & \textcolor{red}{319}                                                            \\ \cline{1-2} \cline{4-7} \cline{9-12} 
\nop & distance            &                                                                & 6                                                                        & \textcolor{red}{63}                                                            & \textcolor{red}{266} (118)                                                     & \textcolor{red}{272}                                                            &                                                                & 12                                                                       & \textcolor{blue}{60}                                                            & \textcolor{red}{204} (123)                                                     & \textcolor{blue}{373}                                                            \\ \cline{1-2} \cline{4-7} \cline{9-12} 
\nop & entropy             &                                                                & 1                                                                        & 50                                                            & 25.2 (130)                                                    & 345                                                            &                                                                & 2                                                                        & 47                                                            & 124 (116)                                                     & 388                                                            \\ \hline 
 
B-\vamp     & N/A                 &   \multirow{4}{*}{3}                                                             & N/A                                                                      & 3.3                                                           & -337 (70)                                                     & 561                                                            &             \multirow{4}{*}{4}                                                   & N/A                                                                      & 0                                                             & -243 (56)                                                     & 750                                                            \\ \cline{1-2} \cline{4-7} \cline{9-12} 
R-\pomcp     & N/A                 &                                                                & N/A                                                                      & 7                                                             & -158 (62.1)                                                     & 665                                                            &                                                                & N/A                                                                      & 0                                                            & -117 (28)                                                     & 764                                                            \\ \cline{1-2} \cline{4-7} \cline{9-12} 
\nop & uniform             &                                      & 15                                                                       & \textcolor{blue}{63}                                                            & \textcolor{blue}{297} (112)                                                     & \textcolor{blue}{463}                                                            &                                       & 24                                                                       & \textcolor{red}{67}                                                            & \textcolor{red}{353} (101)                                                     & \textcolor{red}{504}                                                            \\ \cline{1-2} \cline{4-7} \cline{9-12} 
\nop & distance            &                                                                & 17                                                                       & \textcolor{red}{67}                                                            & \textcolor{red}{367} (96.6)                                                    & \textcolor{red}{452}                                                            &                                                                & 18                                                                       & \textcolor{blue}{50}                                                            & \textcolor{blue}{237} (97)                                                      & \textcolor{blue}{505}                                                            \\ \cline{1-2} \cline{4-7} \cline{9-12} 
\nop & entropy             &                                                                & 4                                                                        & 43                                                            & 132 (105)                                                     & 529                                                            &                                                                & 5                                                                        & 33                                                            & 61.6 (97)                                                     & 560                                                            \\ \hline
\end{tabular}}
\end{table}
\nop is also robust to performance degradation from the underlying reference policies. We ran 4 variants of Random3D with an increasing number of obstacles. The results in \tref{tab:maze3dresults} indicate that \nop's performance (especially with uniform random sampling heuristics) does not degrade much even when the reference policy's (i.e. \rrtc) fail percentage increases due to increasingly narrower passages in the maze. This is because \nop uses the results of \rrtc only to provide a set of alternative sequences of actions to perform. As long as there is sufficient diversity of macro-actions (\ie{} sufficient support of the reference policy), the reference-based \pomdp planner can converge to a reasonable policy of the \pomdp problem fast.  

\begin{table}[!htbp]
\centering
\caption{Results on Multi Drones Tag}
\label{tab:dronecaptureresults}
\begin{tabular}{|c|c|c|c|c|}
\hline
                    & \textbf{Episodes} & \textbf{\begin{tabular}[c]{@{}l@{}}Succ.\\ \%\end{tabular}} & \textbf{\begin{tabular}[c]{@{}l@{}}Acc.\\ Rewards\end{tabular}} & \textbf{Steps}             \\ \hline
B-\vamp                 & N/A               &  7.5                                  &  -1.43 (25)                                &  390 \\ \hline
R-\pomcp             &      99       &      {\color[HTML]{3531FF}20}                                                  &                 {\color[HTML]{3531FF}64 (38)}                                      &       {\color[HTML]{3531FF}345}                  \\ \hline
\nop & 106               & {\color[HTML]{FE0000} 89}                                  & {\color[HTML]{FE0000} 423 (29)}                               & {\color[HTML]{FE0000} 171}  \\ \hline
\end{tabular}
\end{table}
For the high-dimensional planning problem---Multi-Drone Tag, \nop performs at least 4 times better than the rest of the methods as it is the only method that exhibits strategies where drones actively discover and surround the tag. Both B-\vamp and \pomcp cannot easily adapt to the uncertainties from deterministic planning. B-\vamp does not incorporate any uncertainty in the effects of actions and \pomcp fixes a set of macro actions from the reference policy for each belief node based only on a \emph{single} sampled particle.
Hence, the expanded set could be insufficient to fully represent the support of the belief, which is a problem that can be further compounded if the reference policy keeps failing.

\begin{table}[!htbp]
\centering
\caption{Ablation Study of $\epsilon$-Exploration for \nop in Maze2D}
\label{tab: exploration ablation}
\resizebox{0.7\textwidth}{!}{
\begin{tabular}{|c|c|c|c|c|c|c|c|}
\hline
$\epsilon$ & \textbf{\begin{tabular}[c]{@{}c@{}}success\\ \%\end{tabular}} & \textbf{\begin{tabular}[c]{@{}c@{}}E[Tot.Reward]\\ (stdErr)\end{tabular}} & \textbf{\#Steps} & $\epsilon$ & \textbf{\begin{tabular}[c]{@{}c@{}}success\\ \%\end{tabular}} & 
\textbf{\begin{tabular}[c]{@{}c@{}}E[Tot.Reward]\\ (stdErr)\end{tabular}} & \textbf{\#Steps} \\ \hline
0                                & 100                                                        & 761 (2.9)       & 293            & 0.1                              & 87                                                         & 603 (79)        & 378            \\ \hline
0.2                              & 95                                                         & 699 (44)        & 314            & 0.3                              & 87                                                         & 573 (83)        & 374            \\ \hline
0.4                              & 87                                                         & 579 (75)        & 444            & 0.5                              & 80                                                         & 503 (82)        & 460            \\ \hline 
\end{tabular}}
\end{table}
One may be concerned that limiting sub-goal sampling to only the set of states where good observations or rewards can be gained is too restrictive. Therefore, we also evaluate \nop when the sub-goals are sampled in an $\epsilon$-greedy fashion, where with probability $\epsilon$ \nop samples sub-goals from the entire state space and, with $(1-\epsilon)$ probability, it follows the dynamic entropy sampling heuristic mentioned in Section 4.
We evaluated this $\epsilon$-greedy sampling strategy on the Maze2D scenario \tref{tab: exploration ablation}. In this scenario, \nop performs better with lower $\epsilon$. However, as $\epsilon$ increases, its performance does not drop by much.

%% file: sections/discussion.tex
This paper presents a new online approximate \pomdp solver, \nopLong (\nop), which uses \vamp to sample the state space and rapidly generate a large number of macro actions online. These macro-actions reduce the effective planning horizon required and are adaptive to geometric features of the robot's free space. Since reference-based \pomdp planners use macro actions only to bias belief-space sampling and do not require exhaustive enumeration of macro actions, \nop can efficiently exploit many diverse macro actions to compute good \pomdp policies fast. Evaluations on various long horizon \pomdps indicate that \nop outperforms state-of-the-art methods by multiple factors.